\pdfoutput=1

\documentclass[11pt]{article}

\usepackage[preprint]{coling}

\usepackage{booktabs}

\usepackage{times}
\usepackage{latexsym}

\usepackage[T1]{fontenc}

\usepackage[utf8]{inputenc}

\usepackage{microtype}

\usepackage{inconsolata}

\usepackage{graphicx}
\usepackage{todonotes}

\usepackage{caption}
\usepackage{subcaption}

\usepackage{comment}

\usepackage{booktabs}
\usepackage[normalem]{ulem}
\usepackage{adjustbox}
\useunder{\uline}{\ul}{}

\title{Ranking Over Scoring: Towards Reliable and Robust Automated Evaluation of LLM-Generated Medical Explanatory Arguments}

\author{
 \textbf{Iker De la Iglesia\textsuperscript{1}\footnotemark[1]},
 \textbf{Iakes Goenaga\textsuperscript{1}\thanks{\, Equal Contribution.}},
 \textbf{Johanna Ramirez-Romero\textsuperscript{2}},
\\
 \textbf{Jose Maria Villa-Gonzalez\textsuperscript{2}},
 \textbf{Josu Goikoetxea\textsuperscript{1}},
 \textbf{Ander Barrena\textsuperscript{1}}
\\
\\
 \textsuperscript{1}HiTZ Center - Ixa, University of the Basque Country UPV/EHU,
 \\
 \textsuperscript{2}Cruces University Hospital, (Barakaldo, Biscay, Spain)
\\
 \small{
   \{iker.delaiglesia, iakes.goenaga, ander.barrena\}@ehu.eus 
 }
}

\begin{document}
\maketitle
\begin{abstract}
Evaluating LLM-generated text has become a key challenge, especially in domain-specific contexts like the medical field. 
This work introduces a novel evaluation methodology for LLM-generated medical explanatory arguments, relying on Proxy Tasks and rankings to closely align results with human evaluation criteria, overcoming the biases typically seen in LLMs used as judges. 
We demonstrate that the proposed evaluators are robust against adversarial attacks, including the assessment of non-argumentative text. Additionally, the human-crafted arguments needed to train the evaluators are minimized to just one example per Proxy Task.
By examining multiple LLM-generated arguments, we establish a methodology for determining whether a Proxy Task is suitable for evaluating LLM-generated medical explanatory arguments, requiring only five examples and two human experts. 
The Proxy Tasks, LM evaluators, and the code will be made available for reproducibility\footnote{Upon acceptance.}.
\end{abstract}

\section{Introduction}\label{introduction}

\begin{figure}[t!]
    \centering
    \includegraphics[width=\linewidth]{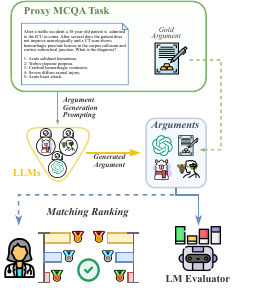}
    \caption{Graphical abstract illustrating the key elements of our approach.
    Synthetic arguments are first generated by prompting multiple LLMs, which are then ranked alongside gold-standard arguments by both our trained LM evaluator and a human expert. Our results show the LM evaluator aligns with human preferences.}
    \label{fig:graphical-abstract}
\end{figure}

The field of Natural Language Processing (NLP) has undergone a transformative evolution with the advent of Language Models (LMs) and Large Language Models (LLMs). The results in the medical domain have been particularly notable, with LLMs achieving remarkable accuracy in solving medical exams \citep{medpalm2, JAMAStanfordUSMLE,ComparativeUSMLE}. This success is driving the ongoing development of these models to further enhance support for Evidence-Based Medicine (EBM) which involves the conscientious, explicit, and thoughtful use of present best medical evidence in making medical decisions \citep{Sackett71}. 
With the advent of large autoregressive generative models, decoder-only architectures such as GPT \cite{GT} and Llama \cite{llama} have been increasingly used for pre-training on medical text data, leading to notable improvements in the coherence and relevance of generated medical explanations. However, the evaluation of such explanatory arguments remains a considerable challenge \citep{chang2023surveyevaluationlargelanguage}, particularly within the medical domain, where obtaining meaningful datasets and assessing accuracy is inherently difficult. The high-entropy nature of language allows for multiple valid responses, complicating the evaluation of relevance, coherence, and factual accuracy. This complexity is further exacerbated by the challenge of objectively quantifying these factors while also accounting for human preferences.

Despite the long tradition of automatic long-text evaluation, particularly in Machine Translation, it remains an unresolved challenge. Metrics like BLEU for translation \cite{bleu}, ROUGE for summarization \citep{rouge}, and embedding-based scores such as BERTScore \citep{bert-score}, BLEURT \citep{sellam2020bleurt}, and COMET \citep{rei-etal-2020-comet} have been widely used, but they present major issues for evaluating explanatory arguments. These metrics rely on reference texts, which are difficult to obtain in the medical domain, and often overestimate irrelevant differences due to the high entropy of valid arguments. This problem extends beyond explanatory argumentation to all long-text evaluations, as noted in the literature \citep{Critics2TraditionalTextEval2023, BLEU-Critic-sulem2018bleu, BertScorCritic-2022bertscore}. 

More recent approaches have tried to address these problems using LLMs as Judges \citep{llmasjudge1,llmasjudge2,llmasjudge3,llmasjudge4}. Despite the increasing popularity of such evaluators, it has been observed that these models often exhibit their own biases; \textit{self-enhancement bias}, tending to recognize and favor their own outputs over those generated by other models, \textit{positional bias}, namely, the propensity to favor certain positions over other and \textit{verbosity bias}, prioritizing lengthier, more verbose responses, even when they fall short in clarity, quality, or accuracy compared to more concise options \citep{bias,llmasjudge1}.

Given this perspective, it is clear that the most reliable way to evaluate the responses generated by LLMs is through human evaluation. Nonetheless, this method has significant drawbacks especially when it comes to highly specialized domains like the medical domain. Finding experts capable of accurately evaluating the responses is very difficult. Added to this, human evaluation is very expensive and time-consuming and in the case of long-text evaluation it is difficult to properly assess quality guidelines. This is particularly evident in extreme cases, where multiple correct responses make the differences too subtle to evaluate, or when the generated texts are incorrect, making it challenging to objectively assess the results. For example, in cases where two argumentative explanations are using different but completely non-sense evidence. 

Considering that explanatory arguments are intended to assist medical decision-making within an EBM framework, the present proposal aims to measure the adequacy of the explanatory arguments in making medical decisions, modeled using proxy tasks along the lines of \cite{proxyqa-ACL2024}. This way, the following \textit{Research Questions} arise:

\begin{itemize}
    \item [\textbf{RQ1}] Can we develop an LM Evaluator that aligns with human preferences when assessing medical explanatory arguments in EBM while avoiding generative LLM Judges bias? 
    \item [\textbf{RQ2}] Are LM evaluators built upon Proxy Tasks that model EBM suitable for evaluating the goodness of LLM-generated medical arguments and robust against adversarial attacks?
    \item [\textbf{RQ3}] Are all Proxy Tasks equally valuable for evaluation purposes?
    \item [\textbf{RQ4}] Are human evaluations consistent across different Proxy Tasks?
    \item [\textbf{RQ5}] Does the LM evaluator with higher task scores produce a ranking of arguments that better aligns with human judgment compared to an LM evaluator with lower task scores?
\end{itemize}

By exploring these research questions, we aim to introduce a fast and cost-effective automatic evaluation method to evaluate medical explanatory argumentation provided by LLMs in the framework of EBM. To do so, the proposed discriminative LM evaluator rather than evaluating the outcome against a reference gold-standard text, evaluates how helpful and informative the generated text is for making medical decisions modeled as three Proxy Tasks, Medical Question Answering, Misinformation, and Natural Language Inference in clinical trials. Besides, this evaluation method aligns with the assessments conducted by expert physicians. We also analyze the adequacy of each Proxy Task for accurate explanatory arguments evaluation. All in all, our proposal eliminates the need for subject matter experts' evaluations as well as the existence of a reference gold-standard explanatory argument while also minimizing some biases of generative LLM Judges.

\section{Related Work}

The development of LLMs in the medical domain focuses nowadays on scaling up pre-training data and model parameters or adapting general-purpose LLMs to the medical domain. Notable examples include Med-PaLM 2 and Meditron. Med-PaLM 2, achieved 86.5\% accuracy on MedQA (US Medical Licensing Exam-style questions), surpassing the previous state-of-the-art \cite{medpalm2}, while Meditron integrates diverse medical information for comprehensive insights and high-quality medical argumentation \cite{meditron}. However, we will not focus on directly assessing the capability of LLMs to solve tasks but rather evaluating the informativeness of LLM-generated explanatory arguments in the medical domain, which is doubly challenging. 

To address the issue of long-text evaluation in a general domain, \citet{proxyqa-ACL2024} propose using a QA task as a proxy to assess the helpfulness and relevance of content generated by LLMs. Their system comprises two key components: meta-questions and proxy-questions. Meta-questions prompt LLMs to generate comprehensive, factually correct text requiring a full understanding of the topic, while proxy-questions evaluate the quality of the generated content by assessing whether it includes sufficient relevant and accurate information. For example, if the meta-question asks about the First Industrial Revolution, a proxy-question might be, "True or False: The steam engine played a crucial role in the First Industrial Revolution." The authors compare their Proxy-QA evaluator with human evaluators and LLM-as-judges, using GPT-as-Judge. They randomly sample ten meta-questions and use four LLMs to generate long-texts. These 4 generated text candidates are then evaluated through pairwise comparisons between three evaluators (their Proxy-QA evaluator, human evaluators, and GPT-as-judge) using the win rate measure\footnote{Win Rate Calculation: The win rate is calculated based on pairwise comparisons of the reports. If one model’s output is preferred over another, it wins that comparison. This win rate measures how often one model’s report is rated better than others by the evaluators.}.

The primary goal of this pairwise comparison in win rate is to determine how closely ProxyQA correlates with human judgment compared to LLM-based evaluations.

The authors found that ProxyQA's evaluations were highly correlated with human preferences, whereas GPT-as-judge tended to overestimate the quality of the text generated by GPT models. ProxyQA showed a balanced and reliable evaluation, reflecting human preferences more closely than GPT evaluators, which were biased towards outputs from GPT-based models. \textcolor{black}{Scalability and domain adaptation is one of the main pitfalls of this method, creating and maintaining high-quality meta-questions is human-intensive. Additionally, the results lack of comparison between the performance of the systems with or without the generated long-text making it difficult to assess the real impact that adding the generated long-text has on solving the Proxy Task.  Our proposal does not require building meta-questions and we include a Naive version of every system where there is no explanatory argument included to solve the Proxy Tasks. We also extend the number of Proxy Tasks.}

\citeauthor{yao2023human}'s work also explores long-text evaluation, with a particular emphasis on human-annotated natural language explanations to assess whether they consistently enhance machine learning models in NLP. Especially relevant for this work is their analysis of how human-annotated explanations show varying levels of helpfulness, depending on the task and dataset used. The study evaluates five large-scale datasets (e.g., CoS-E, e-SNLI) using two NLP models (T5 and BART) to assess explanation quality. Their findings show that explanations in ECQA are highly beneficial, while CoS-E explanations, although noisy, still offer improvements in model predictions. This suggests that explanation evaluation should focus on task-specific performance rather than treating all explanations as equally valuable. \textcolor{black}{While they introduce a metric to assess the helpfulness of long texts, they neither compare different explanations nor verify if their metric aligns with human preferences}.

To summarize, there is an urgent need for an objective system for independent evaluation of modern LLMs’ medical argument generation abilities. To address this, we have developed a medical argumentation evaluation method based on Proxy Tasks that aligns with the assessments of medical experts. Our evaluation method allows us to assess medical argumentations quickly, efficiently, and cost-effectively.

\section{Experimental Setup}

In this study, we developed an experimental framework to investigate the alignment between LM evaluator systems and human preferences in assessing explanatory arguments. 
Argument quality is indirectly estimated by its impact on Proxy Task performance.
These tasks are handled by LMs trained to perform the original task. These LMs also serve as evaluators when incorporating explanatory arguments as additional input, by ranking the incorporated arguments based on the task score. 

The departure point of our approach is the generation of explanatory arguments. Indeed, they comprise the base of our approach, and they will be generated by humans or LLMs. On the one hand, each task will have high-quality arguments written by human experts that we will consider as the gold standard. On the other hand, we will generate diverse arguments for each task using different LLMs. The main focus of the evaluation approach presented in this paper is focused on these two kinds of arguments, termed \textit{Primary Arguments}.

Regarding the Proxy Tasks, we employ a diverse set, including Medical Multiple Choice Question Answering (MMCQA), Medical Misinformation Detection, and Natural Language Inference (NLI) in clinical trials. 
These tasks are selected because they represent different contexts where explanatory argumentation is helpful, each task requiring specific types of arguments. 
By employing a diverse set of Proxy Tasks rather than relying on a single one, we aim to explore which tasks are most relevant and suitable for evaluation purposes (addressing RQs 2 and 3).

We will also have two types of evaluators: human evaluators and LM ones. For the latter, we train discriminative LMs on the Proxy Tasks to function as evaluators, as mentioned previously.
The evaluators thus provide an indirect assessment by leveraging task performance metrics to differentiate between arguments, thereby addressing the potential biases associated with LLMs as Judges-based evaluation methods (RQ1 and RQ2).
Alongside these Proxy Tasks, human experts independently estimate the quality of arguments within the context of these Proxy Tasks, providing a standard against which the evaluators can be compared, addressing RQs 3 and 4.

Therefore, we will have human and LM evaluators, and, essentially, the core of our analysis focuses on examining which of the latter aligns most closely with the former.
We analyze the degree to which the rankings generated by the LM evaluators reflect human preferences, thereby assessing not just task performance but also the meaningfulness of the rankings in the context of human-aligned argument evaluation.  
We also examine whether the LM evaluator with maximized overall Proxy Task score is the one with the closest ranking alignment with human criteria (RQ5).

To further test the robustness and ability to discern the quality of the arguments of our LM evaluators, we introduce a second set of arguments, termed \textit{Control Cases}, which complement \textit{Primary Arguments}. 
Through this approach, LM evaluators are tested with four adversarial scenarios during inference, designed to assess their ability to distinguish meaningful arguments from irrelevant or misleading content, as detailed in \autoref{sec:control-cases}.

Through this experimental setup, we aim to thoroughly investigate the effectiveness of Proxy Task-based evaluators in modeling human judgment and the relative value of different Proxy Tasks.

\subsection{Proxy Tasks \& Proxy Task LM Evaluators}

\subsubsection{Proxy Tasks Benchmarks}\label{tasks}

We repurposed three diverse benchmarks as Proxy Tasks, each selected to capture distinct types of argumentation, offering a broad evaluation across a range of complex scenarios. While all the datasets selected for the tasks include a complementary gold-standard argument supporting the correct label, this argument is unnecessary to perform the base task. Detailed examples of instances from each dataset can be found in \autoref{sec:appendix}.

\paragraph{Medical Multiple Choice QA Benchmark}
We employed the English translation of the \textit{CasiMedicos} dataset \citep{explanatory}, which assesses models' ability to answer medical multiple-choice questions. Each instance includes a question with a clinical case, possible answers, and a gold-standard explanation supporting the correct choice. The original split distribution was kept. To reduce label prior bias, ensuring the model predicts correct answers based on content rather than answer order, we preprocessed the dataset by creating multiple versions of each instance, varying the position of the correct answer. Additionally, we modified the gold-standard explanations by removing statements that explicitly identified the correct answer and replaced references to the answer’s position with the answer's text.

\paragraph{Misinformation Detection Benchmark} 
We employed a subset of the English version of the \textit{HealthFC} dataset \cite{missinformation_detection}, which focuses on indicating whether health-related claims are supported, refuted, or lack enough information.
The dataset contains 742 instances, which we stratified and split into 70\% (518) for training, 15\% (111) for development, and 15\% (112) for testing.
As mentioned in \autoref{sec:human-ranking}, instances labeled ``Not Enough Evidence'' in the test split were excluded from both human and automatic rankings when calculating the final scores. We termed this subset the \textit{Misinformation With Evidence dataset}.

\paragraph{NLI Benchmark}
The \textit{NLI4CT} clinical trial dataset \citep{nli4ct} contains clinical trial records (CTR), including a medical statement and a label indicating whether the CTR supports or contradicts the statement. 
Unlike the other tasks, the arguments in this case are extracted directly from the CTR.
While the original dataset incorporates instances involving two clinical trials, we focused solely on those involving a single trial. 
The distribution is 1035 instances (74\%) for training, 140 instances (10\%) for development, and 229 instances (16\%) for testing.

\subsubsection{Proxy Task LM Evaluators}\label{students}

This section introduces our key contribution: the explanatory argument LM evaluators, designed to systematically rank medical explanatory arguments without direct human involvement. These evaluators will be compared with expert assessments to see which approach aligns most closely with human judgments.
Our method uses discriminative language models trained on Proxy Tasks, avoiding the bias that generative LLMs introduce when acting as evaluators. Generative models tend to favor arguments similar to those they generate, whereas discriminative models focus purely on task performance. This ensures a more objective ranking based on how effectively the arguments improve Proxy Task outcomes.

We developed three evaluators, all based on the EriBERTa encoder model \cite{eriberta}, each trained with different types of arguments.
\autoref{tab:documentexample_IN} in \autoref{sec:appendix_IN} outlines the training inputs for each evaluator based on each Proxy Task.
We used the \textit{train} and \textit{dev} splits for training and tuning, and the \textit{test} split for the final ranking.

\paragraph{Baseline Evaluator}
It serves as the simplest model in this work. It is the original classification task, which means, training without the complementary arguments.

\paragraph{Expert-Trained Evaluator}
Trained using human-crafted gold-standard arguments. This evaluator is expected to align most closely with human judgment, as the training data comes directly from domain experts.

\paragraph{LLM-Trained Evaluator}
One key contribution of this study, an LM trained exclusively with synthetic arguments generated by LLMs (detailed in \autoref{sec:synthetic-arguments}). 
Each training instance includes an argument randomly selected from various LLMs, ensuring a balanced representation.
This approach allows the evaluator to learn diverse argument styles, reducing favoritism toward any specific LLM-generated argument and improving its neutrality and robustness in assessing argument quality.
We trained three models with different argument sets to minimize bias and variability.

\subsection{Primary Arguments and Control Cases}

\begin{figure}[h!]
    \centering
    \includegraphics[width=\linewidth]{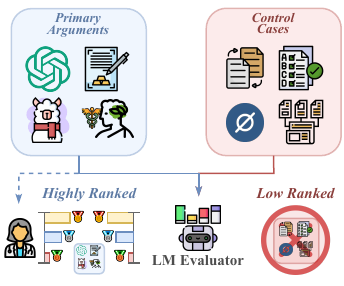}
    \caption{A graphical abstract illustrating the system’s main components and behavior. The proposed LM evaluator prioritizes ranking primary arguments first and placing control cases last.}
    \label{fig:primary-vs-control}
\end{figure}

\textit{Primary Arguments} and \textit{Control Cases} are two main components of our evaluation framework. \textit{Primary Arguments} are central to our research and include both gold-standard arguments crafted by domain experts and synthetic arguments generated by various LLMs. These arguments are the only ones also evaluated by human experts, providing a benchmark for comparing the performance of our automated LM evaluators. In contrast, \textit{Control Cases} are designed to test the robustness of the LM evaluators by incorporating misleading or irrelevant content (see \autoref{fig:primary-vs-control}).

\subsubsection{LLM-Generated Synthetic Arguments} \label{sec:synthetic-arguments}

To evaluate automated medical argumentation, we generated synthetic arguments using three LLMs: GPT-4o \cite{gpt-4o}, known for its strong general reasoning abilities; OpenBioLLM \citep{OpenBioLLMs}, a model fine-tuned on large-scale biomedical datasets for high accuracy in medical text generation; and Llama3-70B-Instruct \citep{Llama3}, which we instruction-tuned for each Proxy Task to optimize its performance in the medical domain (see \autoref{sec:appendix_IT} for an example).

For consistency and minimum human intervention, we used a single example for one-shot prompting for each Proxy Task. 
For MMCQA and Misinformation Detection tasks, the LLMs generated free-style explanatory arguments, while for NLI tasks, they extracted evidence from the CTR. Identical generation parameters were used across all models: maximum token length of 256, sampling enabled, temperature of 0.9, and top-p at 0.85.

\subsubsection{Control Cases}\label{sec:control-cases}

\textit{Control Cases} are designed to assess the LM evaluators' ability to differentiate between meaningful arguments and irrelevant or misleading content. \textit{Control Cases} serve as adversarial attacks for evaluators \cite{jia-liang-2017-adversarial}. This section outlines the construction of these cases and their role in our broader experimental framework. 

\paragraph{No Argument}
This case includes no medical argumentation, testing whether evaluators actually rely on arguments when making predictions. Evaluators trained to evaluate medical arguments are expected to struggle, as they rely on the presence of explanations for predictions.

\paragraph{Label-Only Input}
The correct answers to the Proxy Tasks are provided but without any supporting argumentation. The purpose is to see if evaluators penalize the lack of argumentation, despite having the correct answers. 
We expect evaluators trained on medical argumentation to prioritize explanations and perform worse compared to instances with proper arguments.

\paragraph{Noise Argument}
In this scenario, medical arguments are present but irrelevant to the instance, having been randomly selected from unrelated examples. 
We anticipate that well-trained evaluators will recognize the mismatch and perform poorly, as the arguments do not align with the instance.

\paragraph{IR Passages}
In this test, we use passages from the WikiMed corpus \cite{wikimed}, retrieved via an Information Retrieval (IR) system. While these passages contain medical information, they do not necessarily constitute coherent or valid arguments. This case is designed to challenge evaluators in distinguishing between structured medical arguments and mere informative text. 
Passages were retrieved by indexing full documents with FAISS \cite{faiss} using the \citet{pubmedmarco} embedding model, querying each instance's text, and extracting the top five documents. These were split into 300-character chunks and reranked using ColBERTv2 \cite{colbertv2}, with the top three passages fed to the evaluator.

\subsection{Human and Automatic Ranking}\label{sec:human-ranking}

We engaged two clinicians with prior experience in medical annotation and system evaluation, utilizing the \textit{test} split of the datasets. After a preliminary round, 5 examples were ranked for each task to calculate the Inter-Annotator Agreement (ITA). Experts ranked the four \textit{Primary Arguments} on a scale of 1 to 5, with 5 assigned to clearly incorrect arguments, and ties were allowed when arguments were of equal quality. We used Krippendorff’s alpha \citep{krippendorff} to calculate ITA, achieving the following scores: MMCQA=$0.72$, Misinformation=$0.61$, and NLI=$0.44$.

In the ITA phase, we noticed significant disagreement in the Misinformation Detection task, particularly for instances labeled as ``Not enough evidence''. To address this, we removed those instances and recalculated ITA using 14 new examples, improving the alpha to $0.73$. After this adjustment, the clinicians ranked the arguments independently: 61 instances for MMCQA, 39 for Misinformation, and 98 for NLI.

For both human and LM evaluator rankings, we calculated the average rank for each system. A Friedman non-parametric test \citep{friedman} ($\alpha=5\%$) was applied to assess significant differences in rankings. The NLI task was the only one that failed the Friedman test in the human rankings ($p=0.561$), consistent with its lower ITA.

\begin{figure*}[htb!]
\centering

    \begin{subfigure}{0.32\textwidth}
        \centering
        \includegraphics[width=\textwidth]{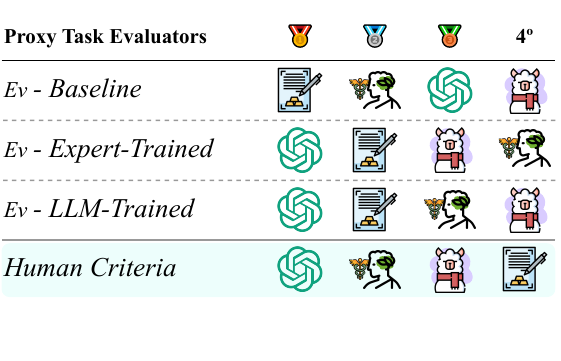}
        \caption{MMCQA}
        \label{fig:ranking:human-criteria:qa}
    \end{subfigure}
    \hfill
    \begin{subfigure}{0.32\textwidth}
        \centering
        \includegraphics[width=\textwidth]{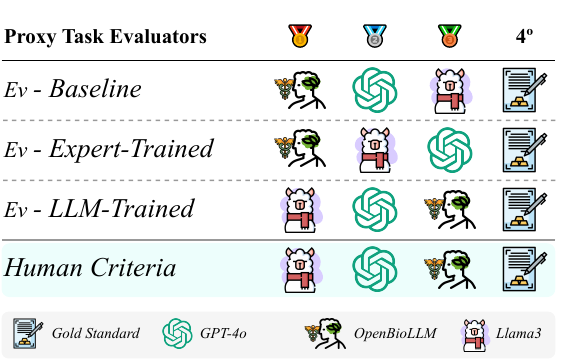}
        \caption{Misinformation With Evidence}
        \label{fig:ranking:human-criteria:misinformation-with-evidence}
    \end{subfigure}    
    \hfill  
    \begin{subfigure}{0.32\textwidth}
        \centering
        \includegraphics[width=\textwidth]{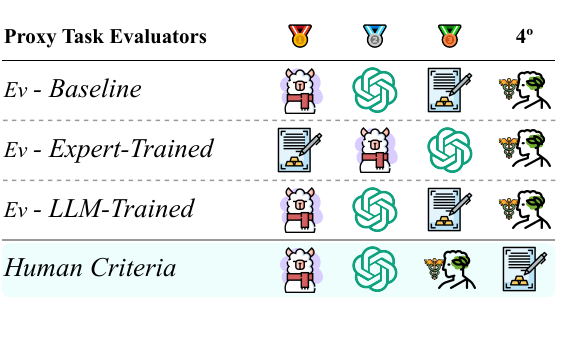}
        \caption{NLI}
        \label{fig:ranking:human-criteria:nli}
    \end{subfigure}
    
    \caption{Ranking of the gold-standard argument alongside those generated by automatic systems. Each row corresponds to a distinct evaluator: the first three rows correspond to our proposed  Proxy Task evaluators based on discriminative classification models, while the last row reflects the human criteria, obtained by having experts directly rank the arguments.}
    \label{fig:ranking:human-criteria}
\end{figure*}

\section{Results}\label{main_results}

This section will first present the main results, comparing the proposed automatic evaluators to human criteria. Finally, we will examine the \textit{Control Cases} to demonstrate the automatic evaluators' ability to discard non-argumentative inputs.

\subsection{Automatic Evaluations Results}\label{automatic_results}

\autoref{fig:ranking:human-criteria} presents the main results of this study. Setting aside evaluator accuracy scores, we focus on the rankings produced by the proposed three discriminative evaluators and human criteria. For the MMCQA task (a), the rankings demonstrate that, in the absence of the gold standard, the LLM-trained evaluator aligns with human criteria when ranking LLM-generated synthetic arguments, with GPT-4o being the top choice in 3 out of 4 rankings. 
\textcolor{black}{The lower ranking of gold-standard arguments by human experts stems from their design for last-year medical students. These arguments prioritize straightforwardness, highlighting only key elements needed to discern the correct answer, assuming prior knowledge. However, in the context of analyzing clinical cases rather than exam preparation, a higher degree of contextualization in the arguments is preferred.}
For the NLI task (c), a similar pattern emerges, but here the finetuned Llama3 model ranks first. 
In the misinformation task (b), the LLM-trained evaluator perfectly matches human criteria, ranking Llama3-generated arguments first.

Regarding the evaluators, the lack of argumentation during training causes the baseline evaluator to produce rankings that do not align with human criteria. In contrast, the expert-trained evaluator improves upon the baseline. However, the LLM-trained approach proves to be the winning strategy, demonstrating that we can effectively evaluate LLM-generated argumentation by using synthetic data and training discriminative evaluators, without relying on human-generated arguments.

As mentioned, LLMs acting as judges tend to overestimate self-generated text and show a preference for longer responses. Our approach addresses the first issue by using an EriBERTa encoder. We also observed that the longest text in the MMCQA task was generated by OpenBioLLM, in the misinformation task by Llama3, and again by OpenBioLLM in the NLI task. The rankings provided by the LLM-trained evaluator in \autoref{fig:ranking:human-criteria} demonstrate that this length bias is absent in our approach for MMCQA and NLI tasks. In the case where the bias appears, such as Llama3 in the misinformation task, human evaluators also ranked it first.

As previously mentioned, the best evaluator does not necessarily produce the highest Proxy Task scores. The left side of \autoref{tab:scoreVsrank} shows the average dataset scores for each evaluator. While the expert-trained evaluator produces the highest scores, the LLM-trained evaluator is the one most aligned with human judgment (see \autoref{fig:ranking:all-systems} and Tables \ref{tab:QA_results}, \ref{tab:MISS_results} and \ref{tab:NLI_results} for details). On the right side of \autoref{tab:scoreVsrank}, when examining the scores of the best system for each evaluator\footnote{The MCQA column represents the scores for the gold-standard, GPT-4o, and GPT-4o. For Misinformation: OpenBioLLM, OpenBioLLM, and Llama3. For NLI: Llama3, the gold standard, and Llama3.}, we observe the same pattern.

\begin{table}[htp]
\begin{adjustbox}{max width=\columnwidth}
\begin{tabular}{@{}lcccccc@{}}
\toprule
                               & \multicolumn{3}{c}{Dataset Average}                       & \multicolumn{3}{c}{Best System Per Evaluator}        \\ \cmidrule(l){2-4} \cmidrule(l){5-7}
\textbf{LM Evaluators} & \textbf{MMCQA}  & \textbf{Misinfo} & \textbf{NLI}   & \textbf{MMCQA}  & \textbf{Misinfo} & \textbf{NLI}   \\ \midrule
\textit{Baseline}              & 36.00          & {\ul 44.56}             & {\ul 61.12}    & 41.17          & 48.30                   & {\ul 61.50}    \\
\textit{Expert Trained}        & \textbf{72.83} & \textbf{58.67}          & \textbf{62.61} & \textbf{82.91} & \textbf{61.22}          & \textbf{67.62} \\
\textit{LLM Trained}           & {\ul 70.85}    & 39.74                   & 58.02          & {\ul 78.90}    & {\ul 49.43}             & 61.12          \\ \bottomrule
\end{tabular}
 \end{adjustbox}
 \caption{The left side shows the average dataset scores for primary arguments (gold-standard and three LLMs), while the right side displays the best system per evaluator for each LM evaluator. The highest score is marked in bold, and the second best is underlined. }
\label{tab:scoreVsrank}
\end{table}

\subsection{Control Cases}\label{automatic_results:control-cases}

\begin{figure}[htb]
\centering
    \begin{subfigure}{\columnwidth}
        \centering
        \includegraphics[width=\columnwidth]{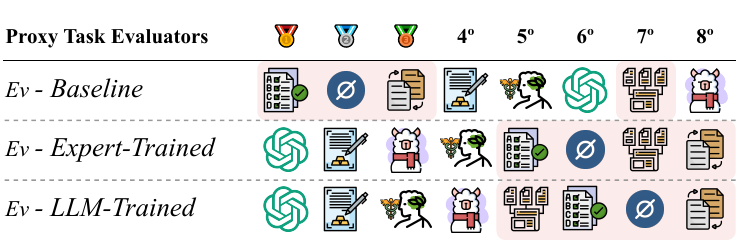}
        \caption{MMCQA}
        \label{fig:ranking:all-systems:qa}
    \end{subfigure}

    \begin{subfigure}{\columnwidth}
        \centering
        \includegraphics[width=\columnwidth]{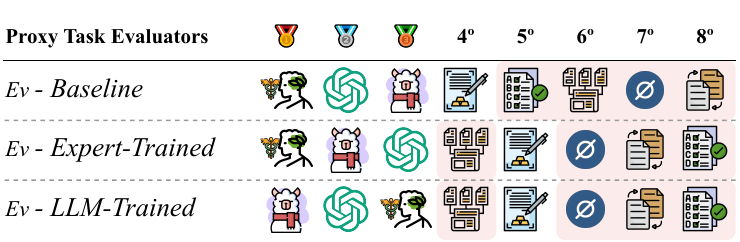}
        \caption{Misinformation With Evidence}
        \label{fig:ranking:all-systems:misinformation-with-evidence}
    \end{subfigure}    

    \begin{subfigure}{\columnwidth}
        \centering
        \includegraphics[width=\columnwidth]{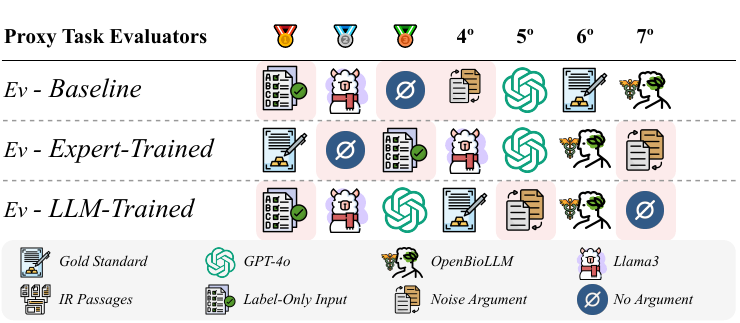}
        \caption{NLI}
        \label{fig:ranking:all-systems:nli}
    \end{subfigure}
    
    \caption{Ranking of the gold-standard argument, LLM-generated arguments, and \textit{Control Cases} by the Proxy Task evaluators for each  Proxy Task. Each row represents a distinct evaluator, and the columns include \textit{Primary Arguments} (gold standard and LLM-generated) as well as \textit{Control Cases} (No Argument, Label-Only Input, Noise Argument, and IR Passages). This table highlights the evaluators' ability to differentiate between proper and improper arguments.}
    \label{fig:ranking:all-systems}
\end{figure}

We have already demonstrated that the LLM-trained evaluator aligns with human criteria when ranking LLM-generated arguments. \autoref{fig:ranking:all-systems} presents an enhanced ranking that includes \textit{Control Cases}, which serve as a form of adversarial attack. Ideally, a robust evaluator should rank all \textit{Control Cases} in the lowest positions. (a) In the MMCQA task, both the Expert and LLM-trained evaluators prefer argumentations over \textit{Control Cases}, while the baseline evaluator is misled by 3 out of 4 \textit{Control Cases}. (b) For the misinformation task, all evaluators perform well, ranking argumentations first and \textit{Control Cases} last. (c) In the NLI task, all models are misled by the \textit{Control Cases}, with the LLM-trained evaluator proving to be the most resilient against control case attacks.

Note that, depending on the task, each control case behaves differently. In MMCQA and NLI, the label-only control case is the most effective attacker, while in the misinformation task, passage retrieval proves to be the strongest. 

\section{Discussion}

Previous work has shown that edit distance-based metrics and LLM-as-a-judge approaches do not serve as reliable evaluators \citep{proxyqa-ACL2024}, highlighting the importance of correlating results with human criteria. In our study, we take this a step further by demonstrating that with less hand-labeled data (one argumentation per Proxy Task, 3 in this work) and a focus on rankings rather than Proxy Task scores (RQ5), we can effectively evaluate LLM generated text and closely match human judgment. We made a deliberate decision not to use a potentially contaminated LLM as an evaluator \cite{conda-2024-data}, opting instead for a discriminative LM evaluator that we confirmed had not been exposed to the datasets used and minimizing some biases of generative LLM Judges (RQ1). 
To our knowledge, we are the first to test \textit{Control Cases}, showing that baseline and expert-trained models are often misled by adversarial attacks while the LLM-trained evaluator remains robust (RQ2). The typical state-of-the-art approaches focus on improving scores on Proxy Tasks assuming expert-trained models as upper bound \cite{ALONSO2024102938}.

Our approach has been validated across three different datasets, demonstrating its robustness and potential for extension to other tasks with minimal human annotations (RQ3). Regarding the human rankings, five examples are sufficient to compute ITA, showing that humans are not consistent across all Proxy Tasks. While there is agreement for MMCQA and misinformation tasks, there is a lack of consensus in the NLI task (clinicians struggle to extract accurate argumentation), indicating that NLI is not suitable for automatic argumentation evaluation (RQ4).

\section{Conclusions}
In this work, we show that across three distinct Proxy Task scenarios, the automatic evaluation of medical explanatory arguments closely aligns with human judgment. Beyond standard MCQA tasks, we broaden our scope to include Misinformation Detection and NLI, providing a more comprehensive assessment.
We present a novel approach that moves beyond traditional score maximization to prioritize improved ranking capabilities, addressing the inherent biases in LLMs when used as judges. Our LLM-trained evaluator aligns closely with human preferences and demonstrates resilience to adversarial attacks. Remarkably, only one hand-labeled example per task is needed to generate the synthetic arguments to develop the LLM-trained evaluator that best resembles human criteria. 
Additionally, we demonstrate that just five examples ranked by two human experts are enough to validate the chosen Proxy Task, confirming the practicality of our evaluation method.

\section{Limitations}
Our approach has the next limitations. First, the discriminative LM model used in this study has a token limit of 512, which may restrict the model’s ability to fully process longer, more complex arguments. However, current advances in expanding language models' context size will mitigate this constraint. Finally, we do not focus explicitly on measuring hallucinations, factual accuracy, or coherence in the generated arguments.

\section*{Acknowledgments}

This work has been partially supported by the HiTZ Center and the Basque Government, Spain (Research group funding IT1570-22). We are also thankful to MCIN/AEI/10.13039/501100011033 projects: (i) Antidote (PCI2020-120717-2), and by European Union NextGenerationEU/PRTR; (ii) DeepKnowledge (PID2021-127777OB-C21) and ERDF A way of making Europe; (iii) EDHIA (PID2022-136522OB-C22). 

\bibliography{custom}

\onecolumn
\appendix

\section{Automatic Evaluator's Inpunts}
\label{sec:appendix_IN}

\begin{table}[h!]
\centering
\scriptsize
\begin{tabular}{l c c c}
\toprule
\textbf{EVALUATORS} & \multicolumn{3}{c}{\textbf{INPUTS}}\\ \midrule 
& \textbf{QA} & \textbf{Missinformation Detection} & \textbf{NLI}
\\ \midrule 
\textbf{Naive Evaluator} & \begin{tabular}[c]{@{}c@{}}Question\\ Clinical Case\\ Possible Answers\\Correct Answer\end{tabular} & \begin{tabular}[c]{@{}c@{}}Question\\ Label\end{tabular} & \begin{tabular}[c]{@{}c@{}}Statement\\ Full Section\\Label\end{tabular}
\\
\midrule
\textbf{Clinician Lined Up Evaluator}  &  \begin{tabular}[c]{@{}c@{}}Question\\ Clinical Case\\ Possible Answers\\\textbf{Gold Argumentation}\\Correct Answer\end{tabular} & \begin{tabular}[c]{@{}c@{}}Question\\ \textbf{Gold Argumentation}\\Label\end{tabular} & \begin{tabular}[c]{@{}c@{}}Statement\\ \textbf{Gold Evidences}\\Label\end{tabular}\\ 
\midrule
\textbf{LLMs Lined Up Evaluators}  &  \begin{tabular}[c]{@{}c@{}}Question\\ Clinical Case\\ Possible Answers\\\textbf{LLMs Argumentation}\\Correct Answer\end{tabular} & \begin{tabular}[c]{@{}c@{}}Question\\ \textbf{LLMs Argumentation}\\Label\end{tabular} & \begin{tabular}[c]{@{}c@{}}Statement\\ \textbf{LLMs Evidences}\\Label\end{tabular}\\ \bottomrule
\end{tabular}
\caption{This table includes the inputs used for each automatic evaluator depending on the proxy task.}
\label{tab:documentexample_IN}
\end{table}

\section{Instruction Tuning Example}
\label{sec:appendix_IT}

\begin{table}[h!]
\centering
\scriptsize
\begin{tabular}{l |p{11.0cm}}
\toprule
\multicolumn{1}{c}{} & \multicolumn{1}{c}{\textbf{Instruction Tuning Example Used For QA}} \\ \midrule 
\textbf{Instruction} & <s>[INST] A 52-year-old man with no concomitant diseases comes to the emergency department for melena of 24 hours evolution without hemodynamic repercussions. He denies taking nonsteroidal anti-inflammatory drugs. The hematocrit is 33\% and the rest of the laboratory tests are normal. The upper gastrointestinal endoscopy performed urgently 6 hours after admission shows a normal stomach, without blood or hematic debris and an excavated ulcer of 8 mm in diameter in the anterior face of the duodenal bulb with "visible vessel" at its base and without active bleeding. Which of the following statements is true? The possible answers are: 1- In the initial endoscopy it is indicated to apply an endoscopic therapy and subsequently to establish endovenous treatment with high doses of a proton pump inhibitor. This strategy has been shown to reduce the risk of hemorrhagic recurrence and mortality. 2- In the initial endoscopy, given the absence of active bleeding, endoscopic therapy is not indicated. Subsequently, to reduce the risk of hemorrhagic recurrence, intravenous treatment with high doses of a proton pump inhibitor should be started. 3- n initial endoscopy, endoscopic therapy is indicated. Subsequent treatment with high doses of a proton pump inhibitor has not been shown to be of any additional benefit. 4- Since this is a complicated ulcer (hemorrhage) the best therapeutic option, once the hemorrhagic episode is resolved, is a vagotomy and pyloroplasty. [/INST]
\\
\midrule
\textbf{Output}  & This is a grade IIa ulcer (Forrest classification), with a high risk of recurrence. Therefore, endoscopic treatment and hospitalization with intravenous treatment with PPIs (it is usually omeprazole) for at least 72 hours is clearly indicated. </s> \\ \bottomrule
\end{tabular}
\caption{This is an example of an instance used to fine-tune Llama3 for QA proxy task applying IT.}
\label{tab:documentexample_IT}
\end{table}

\newpage

\section{Benchmarks Examples}
\label{sec:appendix}

\begin{table}[h!]
\centering
\scriptsize
\begin{tabular}{l |p{11.0cm}}
\toprule
\multicolumn{1}{c}{} & \multicolumn{1}{c}{\textbf{Example of an instance from NLI Benchmark}} \\ \midrule 
\textbf{Statement} &  There were 7 more cases of Anaemia and 1 more case of Disseminated intravascular coagulation in cohort 1 of the primary trial compared to cohort 2. \\ \midrule
\textbf{Gold Evidences}  & Adverse Events 1: **   Total: 158/482 (32.78\%) **   Anaemia 7/482 (1.45\%) **   Disseminated intravascular coagulation 1/482 (0.21\%) ** Adverse Events 2: **   Total: 37/238 (15.55\%) **   Anaemia 2/238 (0.84\%) **   Disseminated intravascular coagulation 0/238 (0.00\%)\\ \midrule
\textbf{Full Document}  & INTERVENTION 1:  **   Everolimus + Exemestane **   Everolimus 10 mg daily in combination with exemestane 25 mg daily ** INTERVENTION 2:  **   Placebo + Exemestane **   Placebo of everolimus in combination with exemestane 25 mg daily ** Inclusion Criteria: **   Adult women ( 18 years of age) with metastatic or locally advanced breast cancer not amenable to curative treatment by surgery or radiotherapy. **   Histological or cytological confirmation of estrogen-receptor positive (ER+) breast cancer **   Postmenopausal women. **   Disease refractory to non steroidal aromatase inhibitors (NSAI), **   Radiological or clinical evidence of recurrence or progression on or after the last systemic therapy prior to randomization. **   Patients must have at least one lesion that can be accurately measured or bone lesions in the absence of measurable disease as defined above. ** Exclusion Criteria: **   HER2-overexpressing patients **   Patients with only non-measurable lesions other than bone metastasis (e.g. pleural effusion, ascites etc.). **   Patients who received more than one chemotherapy line for Advanced Breast Cancer. **   Previous treatment with exemestane or mTOR inhibitors. **   Known hypersensitivity to mTOR inhibitors, e.g. sirolimus (rapamycin). **   Radiotherapy within four weeks prior to randomization **   Currently receiving hormone replacement therapy, **   Other protocol-defined inclusion/exclusion criteria may apply ** Outcome Measurement:  **   Progression-free Survival (PFS) Based on Local Radiology Review of Tumor Assessments. **   Progression-free survival, the primary endpoint in this study, is defined as the time from the date of randomization to the date of first documented radiological progression or death due to any cause. Disease progression was based on the tumor assessment by the local radiologist or investigator using RECIST 1.0 criteria. If a patient did not progress or known to have died at the date of the analysis cut-off or start of another antineoplastic therapy, the PFS date was censored to the date of last adequate tumor assessment prior to cut-off date or start of antineoplastic therapy. For patients with lytic or mixed (lytic+sclerotic) bone lesions, the following is considered progression: appearance of 1 new lytic lesions in bone; the appearance of  new lesions outside of bone and unequivocal progression of existing bone lesions. **   Time frame: date of randomization to the date of first documented tumor progression or death from any cause, whichever occurs first, reported between day of first patient randomized up to about 19 months ** Results 1:  **   Arm/Group Title: Everolimus + Exemestane **   Arm/Group Description: Everolimus 10 mg daily in combination with exemestane 25 mg daily **   Overall Number of Participants Analyzed: 485 **   Median (95\% Confidence Interval) **   Unit of Measure: months  6.93        (6.44 to 8.05) ** Results 2:  **   Arm/Group Title: Placebo + Exemestane **   Arm/Group Description: Placebo of everolimus in combination with exemestane 25 mg daily **   Overall Number of Participants Analyzed: 239 **   Median (95\% Confidence Interval) **   Unit of Measure: months  2.83        (2.76 to 4.14) ** Adverse Events 1: **   Total: 158/482 (32.78\%) **   Anaemia 7/482 (1.45\%) **   Disseminated intravascular coagulation 1/482 (0.21\%) **   Lymphadenopathy 0/482 (0.00\%) **   Neutropenia 0/482 (0.00\%) **   Thrombocytopenia 2/482 (0.41\%) **   Anaemia 28/482 (1.66\%) **   Disseminated intravascular coagulation 21/482 (0.21\%) **   Febrile neutropenia 21/482 (0.21\%) **   Lymphadenopathy 20/482 (0.00\%) **   Neutropenia 20/482 (0.00\%) ** Adverse Events 2: **   Total: 37/238 (15.55\%) **   Anaemia 2/238 (0.84\%) **   Disseminated intravascular coagulation 0/238 (0.00\%) **   Lymphadenopathy 1/238 (0.42\%) **   Neutropenia 1/238 (0.42\%) **   Thrombocytopenia 0/238 (0.00\%) **   Anaemia 22/238 (0.84\%) **   Disseminated intravascular coagulation 20/238 (0.00\%) **   Febrile neutropenia 21/238 (0.42\%) **   Lymphadenopathy 21/238 (0.42\%) **   Neutropenia 21/238 (0.42\%)\\ \midrule
\textbf{Full Section}  & Adverse Events 1: **   Total: 158/482 (32.78\%) **   Anaemia 7/482 (1.45\%) **   Disseminated intravascular coagulation 1/482 (0.21\%) **   Lymphadenopathy 0/482 (0.00\%) **   Neutropenia 0/482 (0.00\%) **   Thrombocytopenia 2/482 (0.41\%) **   Anaemia 28/482 (1.66\%) **   Disseminated intravascular coagulation 21/482 (0.21\%) **   Febrile neutropenia 21/482 (0.21\%) **   Lymphadenopathy 20/482 (0.00\%) **   Neutropenia 20/482 (0.00\%) ** Adverse Events 2: **   Total: 37/238 (15.55\%) **   Anaemia 2/238 (0.84\%) **   Disseminated intravascular coagulation 0/238 (0.00\%) **   Lymphadenopathy 1/238 (0.42\%) **   Neutropenia 1/238 (0.42\%) **   Thrombocytopenia 0/238 (0.00\%) **   Anaemia 22/238 (0.84\%) **   Disseminated intravascular coagulation 20/238 (0.00\%) **   Febrile neutropenia 21/238 (0.42\%) **   Lymphadenopathy 21/238 (0.42\%) **   Neutropenia 21/238 (0.42\%) \\ \midrule

\textbf{Label}  & Entailment\\ \bottomrule
\end{tabular}
\caption{An instance example from NLI Benchmark.}
\label{tab:instanceexample_NLI}
\end{table}

\begin{table}[h!]
\centering
\scriptsize
\begin{tabular}{l |p{11.0cm}}
\toprule
\multicolumn{1}{c}{} & \multicolumn{1}{c}{\textbf{Example of an instance from Missinformation Detection Benchmark}} \\ \midrule 
\textbf{Question} & Can filtering out blue light using blue filter glasses or night mode settings on smartphone, tablet or laptop screens have a beneficial effect on sleep? \\ \midrule
\textbf{Gold Argumentation}  & In previous studies, it makes no noticeable difference to sleep when the blue light component of display screen devices is filtered out in the evening. However, the results are not well validated because the studies are of low quality and usually only examined a few people.\\ \midrule

\textbf{Label}  & Refuted\\ \bottomrule
\end{tabular}
\caption{An instance example from Missinformation Detection Benchmark.}
\label{tab:documentexample_missinformation}
\end{table}

\begin{table}[h!]
\centering
\scriptsize
\begin{tabular}{l |p{11.0cm}}
\toprule
\multicolumn{1}{c}{} & \multicolumn{1}{c}{\textbf{Example of a document from the Preprocessed Antidote CasiMedicos Dataset}} \\ \midrule 
\textbf{C} & A 45-year-old man undergoes a truncal vagotomy and antrectomy with Billroth II reconstruction for chronic peptic ulcer disease with pyloro-duodenal stricture. Six weeks after the surgery she reports that shortly after (less than half an hour) after ingestions she presents nausea, asthenia and sweating, dizziness and abdominal cramps usually accompanied by diarrhea. \\ \midrule
\textbf{Q} & Which of the following is the most appropriate approach for her initial management? \\ \midrule
\textbf{P} & \begin{tabular}[c]{@{}p{11.0cm}}\textbf{(1)} Apply treatment with a somatostatin inhibitor (octreotide). \\ \textbf{(2)} Follow specific dietary measures. \\ \textbf{(3)} Trial treatment with a benzodiazepine.\\ \textbf{(4)} Search for a probable neuroendocrine tumor (e.g. carcinoid). \\ \textbf{(5)} Indicate surgical treatment to perform an antiperistaltic Roux-en-Y gastrojejunostomy.\end{tabular}  \\
\midrule
\textbf{E} & Answers 1, 2 and 5 are appropriate treatments for dumping syndrome or postgastrectomy, but the question is focused on initial management, so the most appropriate answer seems to be 2. \\ 
\midrule
\textbf{NE}  & Applying treatment with a somatostatin inhibitor (octreotide), following specific dietary measures and indicating surgical treatment to perform an antiperistaltic Roux-en-Y gastrojejunostomy are appropriate treatments for dumping syndrome or postgastrectomy, but the question is focused on initial management, so the most appropriate approach seems to be following specific dietary measures.\\ \bottomrule
\end{tabular}
\caption{Example of a document in the Preprocessed Antidote CasiMedicos dataset with the explanation about the correct answer manually neutralized. \textbf{C}: Clinical Case;
\textbf{Q}: Question; \textbf{P}: Possible Answers; \textbf{E}: Correct Answer Explanation. The \emph{Clinical Case}, \emph{Question}, \emph{Possible Answers}, \emph{Correct Answer Explanation}  sections are the original annotations of the Antidote CasiMedicos dataset. The preprocessing of the medical doctors' explanations (\textbf{NE}) is part of this work.}
\label{tab:documentexample}
\end{table}

\newpage

\section{Prompts For Medical Argumentation Generation}
\label{sec:appendix_pompts}

\begin{table}[h!]
\centering
\scriptsize
\begin{tabular}{l |p{11.0cm}}
\toprule
\multicolumn{1}{c}{} & \multicolumn{1}{c}{\textbf{Prompt used to generate medical argumentation for QA}} \\ \midrule 
\textbf{"role": "system", "content":} & You are a medical student and given a medical case, a question and five possible answers, tell me which is the correct answer and argument in favor of it.\newline
Example: 
\newline
A medical case and a question related to it \verb|<casequestion>| After a traffic accident a 38-year-old patient is admitted to the ICU in coma. After several days the patient does not improve neurologically and a CT scan shows hemorrhagic punctate lesions in the corpus callosum and cortico-subcortical junction. What is the diagnosis?\newline
\verb|<\casequestion>|\newline
And five possible answers:\newline
\verb|<ans>|1- Acute subdural hematoma.\verb|<\ans>|\newline
\verb|<ans>|2- Trobocytopenic purpura.\verb|<\ans>| \newline
\verb|<ans>|3- Cerebral hemorrhagic contusion.\verb|<\ans>| \newline
\verb|<ans>|4- Severe diffuse axonal injury.\verb|<\ans>| \newline
\verb|<ans>|5- Acute heart attack.\verb|<\ans>| \newline
The argument for the correct answer without mentioning the options and focusing exclusively on the arguments is:\newline
Diffuse axonal injury produces an early and sustained deterioration of the level of consciousness (as mentioned in the case statement) without a lesion on CT scan to justify the picture. Sometimes, punctate hemorrhages at the level of the corpus callosum, corticosubcortical junction and dorsolateral portion of the brainstem are evidenced in this imaging test. \\
\midrule
\textbf{"role": "user", "content":}  & Given this new case and the question related to it: \newline
\verb|<casequestion>| \verb|{case_question} |\verb|<\casequestion>|\newline
And five possible answers:\newline
\verb|<ans>| \verb|{ans1}| \verb|<\ans>|\newline
\verb|<ans>| \verb|{ans2}| \verb|<\ans>| \newline
\verb|<ans>| \verb|{ans3}| \verb|<\ans>| \newline
\verb|<ans>| \verb|{ans4}| \verb|<\ans>| \newline
\verb|<ans>| \verb|{ans5}| \verb|<\ans>| \newline
The argument for the correct answer without mentioning the options and focusing exclusively on the arguments is: \\ \bottomrule
\end{tabular}
\caption{This is the prompt we used to generate medical argumentation for QA, where \{case\_question\} is a new clinical case and a question related to it from the dataset, and \{ans1-5\} are the possible answer options for the question.  The same prompt has been used on GPT-4o, OpenBioLLM and Llama3.}
\label{tab:documentexample_QA}
\end{table}

\begin{table}[h!]
\centering
\scriptsize
\begin{tabular}{l |p{11.0cm}}
\toprule
\multicolumn{1}{c}{} & \multicolumn{1}{c}{\textbf{Prompt used to generate medical argumentation for Missinformation Detection}} \\ \midrule 
\textbf{"role": "system", "content":} & You are a medical student. Given a medical question, you must answer the question and include the arguments you use to reach your answer.\newline
Example: 
\newline
A question \verb|<question>| Can taking the enzyme diamino oxidase prevent alcohol-related hangover symptoms? \verb|<\question>|\newline
The argument for the correct answer and focusing exclusively on the arguments is:\newline
Such an effect is not likely, nor do clinical studies exist on this issue. \\
\midrule
\textbf{"role": "user", "content":}  & Given this new question: \newline
\verb|<question>| \verb|{question} |\verb|<\question>|\newline
The argument for the correct answer and focusing exclusively on the arguments is: \\ \bottomrule
\end{tabular}
\caption{This is the prompt we used to generate medical argumentation for Missinformation Detection, where \{question\} is a new question from the dataset. The same prompt has been used on GPT-4o, OpenBioLLM and Llama3.}
\label{tab:documentexample_Miss}
\end{table}

\begin{table}[h!]
\centering
\scriptsize
\begin{tabular}{l |p{11.0cm}}
\toprule
\multicolumn{1}{c}{} & \multicolumn{1}{c}{\textbf{Prompt used to extract medical argumentation for NLI}} \\ \midrule 
\textbf{"role": "system", "content":} & You are a medical student. Given a medical hypothesis and evidences separated by **, extract the evidences that supports or contradicts the hypothesis without adding any other words. Remember, do not generate any new text. Extract only the relevant parts exactly as they appear in the given text.\newline
Example: 
\newline
A hypothesis \verb|<hypothesis>| Patients with significantly elevated ejection fraction are excluded from the primary trial, but can still be eligible for the secondary trial if they are 55 years of age or over. \verb|<\hypothesis>| \newline\newline
A list of possible evidences \verb|<evidences>| Inclusion criteria: ** Inclusion Criteria: **   Female patients age 18 years or older **   Histologically proven breast cancer after failure or relapse of no more than three lines of chemotherapy including adjuvant, irrespective of prior hormone therapy metastatic disease (stage IV); **   HER2-negative patients (HER2 1+ or negative, or HER2 2+ and FISH negative) **   At least one measurable tumour lesion (RECIST); ** Exclusion criteria: ** Exclusion Criteria: **   Active infectious disease **   Gastrointestinal disorders that may interfere with the absorption of the study drug or chronic diarrhoea **   Serious illness, concomitant non-oncological disease or mental problems considered by the investigator to be incompatible with the protocol **   Active/symptomatic brain metastases **   Cardiac left ventricular function with resting ejection fraction < 50\% (below upper limit of normal) **   ANC less than 1500/mm3 platelet count less than 100 000/mm3 **   Bilirubin greater than 1.5 mg /dl (>26 and 61549 mol /L, SI unit equivalent) **   AST and ALT greater than 2.5 times the upper limit of normal or greater 5 times the upper limit of normal in case of known liver metastases **   Serum creatinine greater than 1.5 mg/dl (>132 and 61549 mol/L, SI unit equivalent) **   Patients who are sexually active and unwilling to use a medically acceptable method of contraception **   Pregnancy or breast-feeding **   Concomitant treatment with other investigational drugs or other anti-cancer-therapy during this study and/or during the past two/four weeks, prior to the first treatment with the trial drug. Concurrent treatment with biphosphonates is allowed **   Previous treatment with trastuzumab, EGFR-, or EGFR/HER2-inhibitors patients unable to comply with the protocol **   Active alcohol or drug abuse **   Other malignancy within the past 5 years'
'Premenopausal women 55 years of age or younger with regular menstrual cycles (at least four cycles in the last six months). Women with fewer than 4 menses in the last 6 months or who have had a hysterectomy with ovaries intact will be considered premenopausal if FSH level < 20. **   Women with breast density  25\% (scattered fibroglandular densities or greater) are eligible. **   Prior Treatment **   Patients who are currently receiving hormone replacement therapy (estrogen or progesterone); or are taking tamoxifen or raloxifene are not eligible. Women who have taken these medications must have stopped for at least 4 months prior to study entry. **   Topical estrogen (eg, transdermal patches and vaginal estrogens) is allowed. **   Patients with a diagnosis of osteoporosis with physician recommendation for treatment of low bone mass are not eligible. **   Patients known to have hyperparathyroid disease or other serious disturbances of calcium metabolism requiring intervention in the past 5 years are not eligible. **   Patients with a history of kidney stones (unless documented not to have been a calcium stone) are not eligible. **   Patients participating in a concurrent breast cancer chemoprevention trial are not eligible. **   Required initial laboratory values - Calcium < 10.5 mg/dL'   \verb|<\evidences>| \newline
The evidences that supports or contradicts the hypothesis without adding any other words are:
\newline
Cardiac left ventricular function with resting ejection fraction < 50\% (below upper limit of normal). ** Premenopausal women 55 years of age or younger with regular menstrual cycles (at least four cycles in the last six months). \\
\midrule
\textbf{"role": "user", "content":}  & Given this new hypothesis: \newline
\verb|<hypothesis>| \verb|{statement} |\verb|<\hypothesis>|\newline
And given this new list of possible evidences \verb|<evidences>| \verb|{evidences} |\verb|<\evidences>|\newline
The evidences that supports or contradicts the hypothesis without adding any other words are: \\ \bottomrule
\end{tabular}
\caption{This is the prompt we used to extract medical argumentation for NLI, where \{statement\} is a new hypothesis from the dataset and \{evidences\} is a new list of evidences from the dataset related to the hypothesis. The same prompt has been used on GPT-4o, OpenBioLLM and Llama3.}
\label{tab:documentexample_NLI}
\end{table}

\newpage

\section{Experiment Results}

\begin{table}[h!]
\centering
\scriptsize
\begin{tabular}{l c c c c c c c c}
\toprule
\textbf{EVALUATORS\textbackslash TESTS} & \textbf{No Argument} & \textbf{Gold} & \textbf{Noise} & \textbf{Correct} & \textbf{IR} & \textbf{GPT4} & \textbf{OpenBioLLM} & \textbf{Llama3} \\ \midrule 

\textbf{Ev - Baseline} & 40.61 & 37.25 & \textbf{41.17} & 40.62 & 35.01 & 34.73 & 36.97 & 35.01 \\
\textbf{Ev - Expert Trained} & 37.81 & 77.59 & 31.37 & 39.66 & 38.38 & \textbf{82.91} & 64.99 & 65.83 \\
\textbf{Ev - LLM Trained} & 34.45 & 72.64 & 33.79 & 35.89 & 38.39 & \textbf{78.90} & 66.85 & 64.98 \\

\bottomrule
\end{tabular}
\caption{This table includes the accuracy obtained by the different automatic evaluators on selected tests in QA proxy task. Best result for each evaluator in bold.}
\label{tab:QA_results}
\end{table}

\begin{table}[!h]
\centering
\scriptsize
\begin{tabular}{l c c c c c c c c}
\toprule
\textbf{EVALUATORS\textbackslash TESTS}& \textbf{No Argument} & \textbf{Gold} & \textbf{Noise} & \textbf{Correct} & \textbf{IR} & \textbf{GPT4} & \textbf{OpenBioLLM} & \textbf{Llama3}\\ \midrule 

\textbf{Ev - Baseline} & 35.37 & 40.81 & 30.61 & 39.45 & 37.41 & 44.89 & \textbf{48.30} & 44.21 \\
\textbf{Ev - Expert Trained} & 46.25 & 53.06 & 13.60 & 0.68 & 56.46 & 59.86 & \textbf{61.22} & 60.54 \\
\textbf{Ev - LLM Trained} & 18.02 & 25.85 & 11.56 & 5.21 & 38.09 & 42.85 & 40.81 & \textbf{49.43} \\
 \bottomrule
\end{tabular}
\caption{This table includes the accuracy obtained by the different automatic evaluators on selected tests in Missinformation Detection proxy task. Best result for each evaluator in bold.}
\label{tab:MISS_results}
\end{table}

\begin{table}[!h]
\centering
\scriptsize
\begin{tabular}{l c c c c c c c}
\toprule
\textbf{EVALUATORS\textbackslash TESTS}& \textbf{No Argument} & \textbf{Gold} & \textbf{Noise} & \textbf{Correct}  & \textbf{GPT4} & \textbf{OpenBioLLM} & \textbf{Llama3}\\ \midrule 

\textbf{Ev - Baseline} & 60.31 & 60.46 & 60.80 & 62.33 & 60.38 & 60.63 & \textbf{61.50} \\
\textbf{Ev - Expert Trained} & 65.10 & \textbf{67.62} & 60.61 & 64.58 & 61.89 & 61.35 & 63.06 \\
\textbf{Ev - LLM Trained} & 54.32 & 57.34 & 56.15 & \textbf{61.12} & 58.21 & 54.73 & 60.58 \\
 \bottomrule
\end{tabular}
\caption{This table includes the micro F-score obtained by the different automatic evaluators on selected tests in NLI proxy task. Best result for each evaluator in bold.}
\label{tab:NLI_results}
\end{table}

\end{document}